# Image Segmentation Keras : Implementation of Segnet, FCN, UNet, PSPNet and other models in Keras

Divam Gupta
Carnegie Mellon University
divamg@alumni.cmu.edu

## Abstract

*Semantic segmentation plays a vital role in computer vision tasks, enabling precise pixel-level understanding of images. In this paper, we present a comprehensive library for semantic segmentation, which contains implementations of popular segmentation models like SegNet, FCN, UNet, and PSPNet. We also evaluate and compare these models on several datasets, offering researchers and practitioners a powerful toolset for tackling diverse segmentation challenges.*

## 1. Introduction

Semantic segmentation, an essential task in the field of computer vision, aims to assign precise labels to every pixel in an image. Semantic segmentation has wide-ranging applications in autonomous driving, image and video analysis, medical imaging, scene understanding etc. Over the years, deep learning approaches have achieved remarkable success in semantic segmentation.

In this paper, we present a comprehensive library for semantic segmentation [1], aimed at the machine learning community. We also compare various semantic segmentation models on multiple datasets. The library provides an easy to use interface and is built using the TensorFlow and Keras framework. The library offers an extensive collection of easy-to-use models, including SegNet [1], FCN [6], UNet [8], and PSPNet [10], which are widely used networks for semantic segmentation.

The primary objective behind the development of this library is to provide a user-friendly and accessible platform for machine learning researchers and practitioners interested in semantic segmentation. Our library promotes modular and extensible design principles, allowing users to easily integrate, customize, and extend existing segmentation models to meet their specific needs.

---
[1]The code is available at https://github.com/divamgupta/image-segmentation-keras

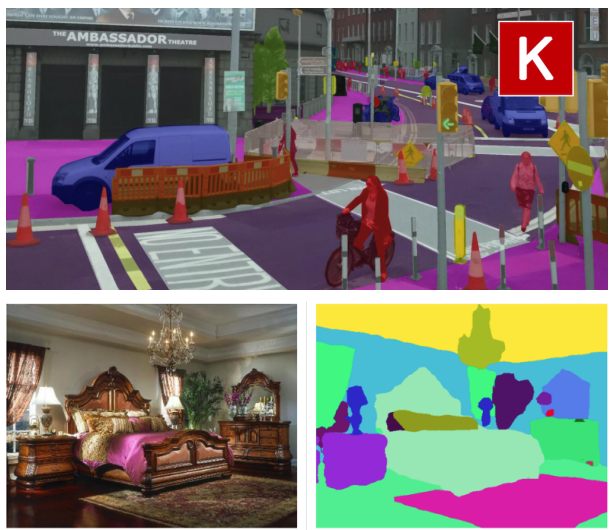

To ensure ease of use, we have extensively documented the library, providing clear instructions and code examples. This documentation includes architectural details and training procedures Additionally, we provide pre-trained weights, enabling users to quickly fine-tune the models on their datasets.

## 2. Networks for Semantic Segmentation

Like most of the other applications, using a CNN for semantic segmentation is the obvious choice. When using a CNN for semantic segmentation, the output is also an image rather than a fixed length vector.

### 2.1. Fully Convolutional Network

Usually, the architecture of the model contains several convolutional layers, non-linear activations, batch normalization, and pooling layers. The initial layers learn the low-level concepts such as edges and colors and the later level layers learn the higher level concepts such as different ob-

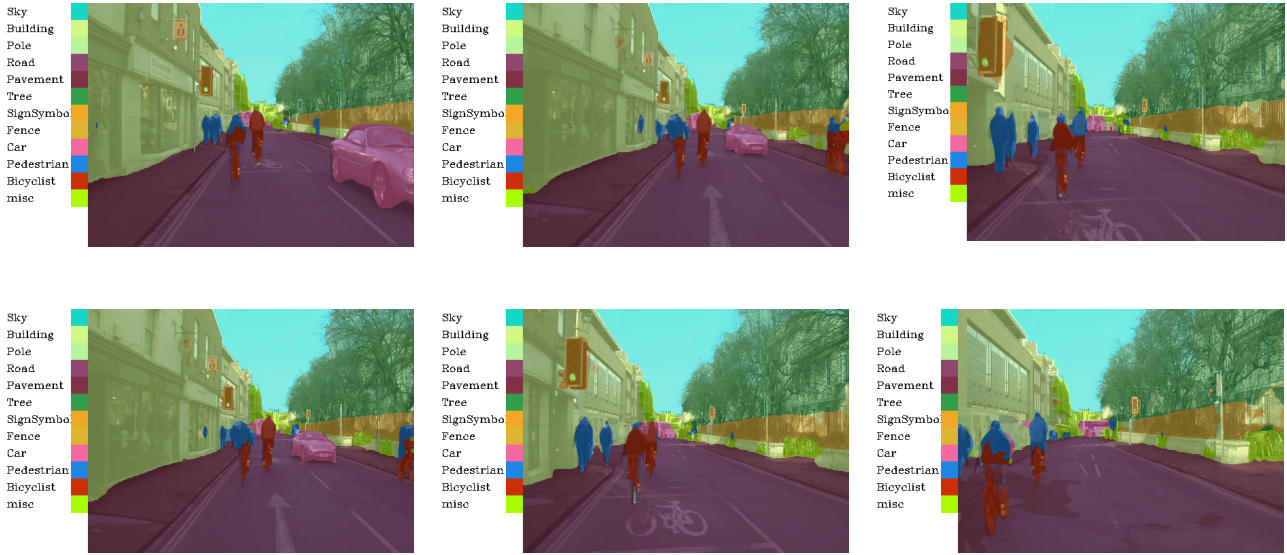

Figure 1. Qualitative results on the CamVid dataset

| Method | mIoU | fIoU | Sky | Buildg | Pole | Road | Pav | Tree | Signl | Fence | Car | Ped | Bicyc | misc |
|---|---|---|---|---|---|---|---|---|---|---|---|---|---|---|
| MobileNet UNet | 64.51 | 86.08 | 94.3 | 84.03 | 12.9 | 96.59 | 84.98 | 88.28 | 32.69 | 52.25 | 85.25 | 42.43 | 72.68 | 27.77 |
| VGG UNet | 59.59 | 84.13 | 93.22 | 81.28 | 9.98 | 95.76 | 81.4 | 89.59 | 33.99 | 57.0 | 69.12 | 30.69 | 50.89 | 22.18 |
| ResNet50 UNet | 60.38 | 85.89 | 94.01 | 85.42 | 8.54 | 96.71 | 86.91 | 89.63 | 44.21 | 65.93 | 55.98 | 38.0 | 33.76 | 25.45 |
| SegNet | 40.89 | 73.4 | 81.1 | 74.31 | 3.14 | 92.13 | 68.46 | 75.05 | 9.19 | 8.61 | 32.06 | 7.32 | 25.79 | 13.53 |
| PSP Pretrained | 65.88 | 86.5 | 93.15 | 86.29 | 12.03 | 94.87 | 81.89 | 90.04 | 51.53 | 71.65 | 81.72 | 37.39 | 59.52 | 30.43 |
| ResNet50 PSPNet | 51.42 | 80.08 | 90.15 | 79.15 | 5.17 | 92.96 | 75.24 | 86.68 | 25.14 | 51.06 | 47.33 | 20.08 | 23.24 | 20.82 |

Table 1. Quantitative results of semantic segmentation on the CamVid dataset.

jects. [2]

At a lower level, the neurons contain information for a small region of the image, whereas at a higher level the neurons contain information for a large region of the image. Thus, as we add more layers, the size of the image keeps on decreasing and the number of channels keeps on increasing. The downsampling is done by the pooling layers.

For the case of image classification, we need to map the spatial tensor from the convolution layers to a fixed length vector. To do that, fully connected layers are used, which destroy all the spatial information. For the task of semantic segmentation, we need to retain the spatial information, hence no fully connected layers are used. That's why they are called fully convolutional networks. The convolutional layers coupled with downsampling layers produce a low-resolution tensor containing the high-level information.

Taking the low-resolution spatial tensor, which contains high-level information, we have to produce high-resolution segmentation outputs. To do that we add more convolution layers coupled with upsampling layers which increase the size of the spatial tensor. As we increase the resolution, we decrease the number of channels as we are getting back to the low-level information.

This is called an encoder-decoder structure. Where the layers which downsample the input are the part of the encoder and the layers which upsample are part of the decoder. When the model is trained for the task of semantic segmentation, the encoder outputs a tensor containing information about the objects, and its shape and size. The decoder takes this information and produces the segmentation maps.

### 2.2. Skip Connections

If we simply stack the encoder and decoder layers, there could be loss of low-level information. Hence, the boundaries in segmentation maps produced by the decoder could be inaccurate.

To make up for the information lost, we let the decoder access the low-level features produced by the encoder layers. That is accomplished by skip connections. Intermediate outputs of the encoder are added/concatenated with the inputs to the intermediate layers of the decoder at appropriate positions. The skip connections from the earlier layers provide the necessary information to the decoder layers which is required for creating accurate boundaries.

---

[2]Source: https://divamgupta.com/image-segmentation/2019/06/06/deep-learning-semantic-segmentation-keras.html

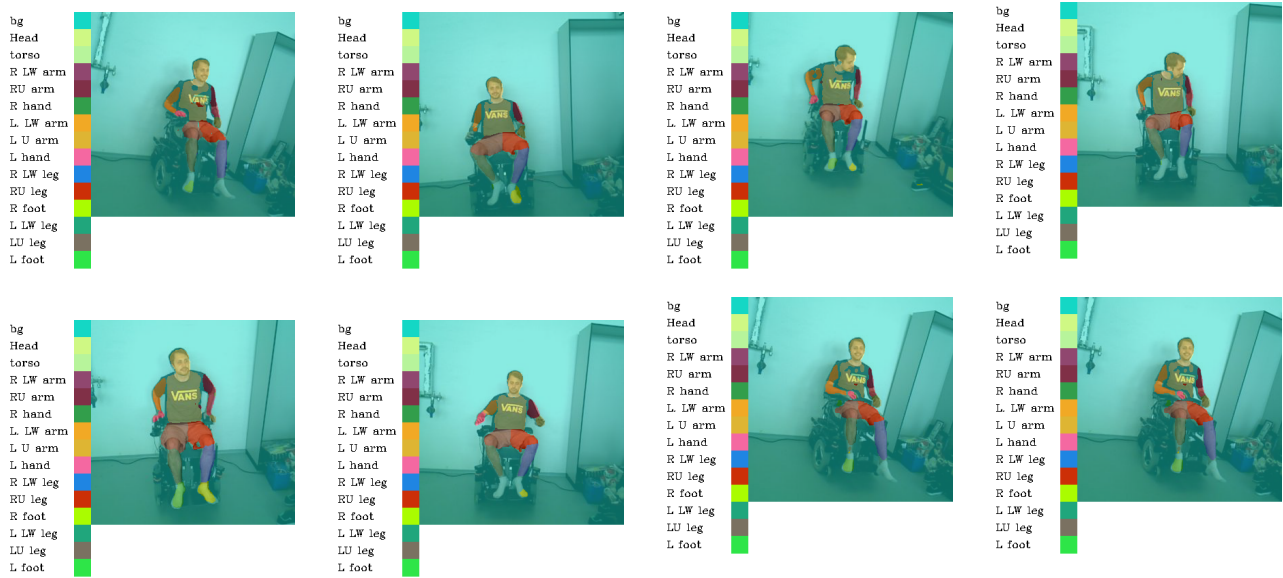

Figure 2. Qualitative results on the sitting people dataset

| Method | mIoU | fIoU | BG | Head | tor | R LW a | RU a | R h | L. LW a | L U a | L h | R LW l | RU l | R f | L LW l | LU l | L f |
|---|---|---|---|---|---|---|---|---|---|---|---|---|---|---|---|---|---|
| VGG UNet | 48.2 | 91.59 | 96.03 | 77.7 | 62.08 | 31.41 | 28.63 | 34.22 | 31.63 | 16.59 | 35.42 | 66.36 | 65.38 | 25.88 | 64.31 | 65.2 | 22.23 |
| PSP Pretrained | 62.19 | 93.95 | 97.02 | 83.18 | 76.42 | 43.5 | 63.2 | 49.05 | 47.1 | 57.39 | 51.15 | 79.15 | 73.54 | 27.82 | 77.05 | 80.94 | 26.31 |
| ResNet50 UNet | 62.16 | 94.2 | 97.25 | 84.28 | 77.04 | 37.36 | 55.47 | 40.14 | 45.42 | 50.7 | 51.5 | 79.36 | 68.84 | 40.6 | 81.58 | 76.37 | 46.46 |
| ResNet50 PSPNet | 44.83 | 91.15 | 95.51 | 77.11 | 74.71 | 12.91 | 43.1 | 18.38 | 18.66 | 42.64 | 22.04 | 52.06 | 64.59 | 24.04 | 45.72 | 64.6 | 16.39 |
| SegNet | 49.26 | 92.48 | 96.62 | 78.24 | 68.49 | 27.44 | 43.07 | 24.39 | 24.62 | 44.17 | 18.93 | 72.71 | 63.22 | 24.93 | 72.21 | 64.11 | 15.81 |
| MobileNet UNet | 58.6 | 93.8 | 97.06 | 78.91 | 80.23 | 34.41 | 43.31 | 44.2 | 38.39 | 52.19 | 49.76 | 72.7 | 68.8 | 30.8 | 77.81 | 76.59 | 33.78 |

Table 2. Quantitative results of semantic segmentation on the sitting people dataset.

## 3. Experiments

In this section we compare various implementations of segmentation models in several datasets.

**CamVid dataset :** CamVid [2] is a unique dataset that provides pixel-level semantic labels for driving scenario videos, with annotations for 11 semantic classes. It offers over 10 minutes of high-quality footage, along with corresponding semantically labeled images, calibration sequences.

**Sitting people dataset :** The Human Part Segmentation dataset [7] by the University of Freiburg is specifically designed for semantic segmentation of sitting people. It comprises various human parts, such as hands, legs, and arms, and contains approximately 15 distinct classes.

**SUIM dataset :** The SUIM (Semantic Underwater Image Manipulation) [5] dataset is a comprehensive collection of underwater imagery. It consists of more than 1500 images with pixel-level annotations for eight object categories, including fish (vertebrates), reefs (invertebrates), aquatic plants, wrecks/ruins, human divers, robots, and sea-floor. These images are gathered during oceanic explorations and human-robot collaborative experiments, and annotated by human participants.

We benchmark the following models:

**SegNet:** Standard encoder-decoder network, where the encoder network produces an input feature map, and decoder predicts the segmentation classes at the input resolution. This has no skip connections or any pretraining.

**MobileNet UNet :** Efficient and accurate semantic segmentation, combining MobileNet's [4] lightweight feature extraction with U-Net's precise pixel-wise predictions. Ideal for real-time applications and resource-constrained environments.

| Method | mIoU | fIoU | Background | Human | Plants | Wreks | Robots | Feefs | Fish | Floor |
|---|---|---|---|---|---|---|---|---|---|---|
| SegNet | 17.03 | 36.7 | 68.13 | 11.94 | 0.0 | 0.0 | 0.0 | 31.04 | 13.8 | 11.3 |
| ResNet50 PSPNet | 15.71 | 33.8 | 63.55 | 10.77 | 0.14 | 11.71 | 0.0 | 25.98 | 4.42 | 9.12 |
| MobileNet UNet | 31.38 | 55.96 | 85.0 | 24.65 | 0.0 | 6.4 | 0.02 | 45.3 | 39.44 | 50.21 |
| VGG UNet | 24.51 | 46.6 | 76.5 | 9.34 | 16.21 | 5.08 | 0.0 | 36.86 | 16.76 | 35.36 |
| ResNet50 UNet | 29.38 | 52.11 | 80.74 | 24.37 | 6.83 | 12.99 | 0.0 | 43.18 | 24.41 | 42.52 |
| PSP Pretrained | 24.03 | 49.21 | 76.41 | 0.22 | 0.0 | 11.99 | 0.0 | 43.9 | 14.11 | 45.64 |

Table 3. Quantitative results of semantic segmentation on the SUIM dataset.

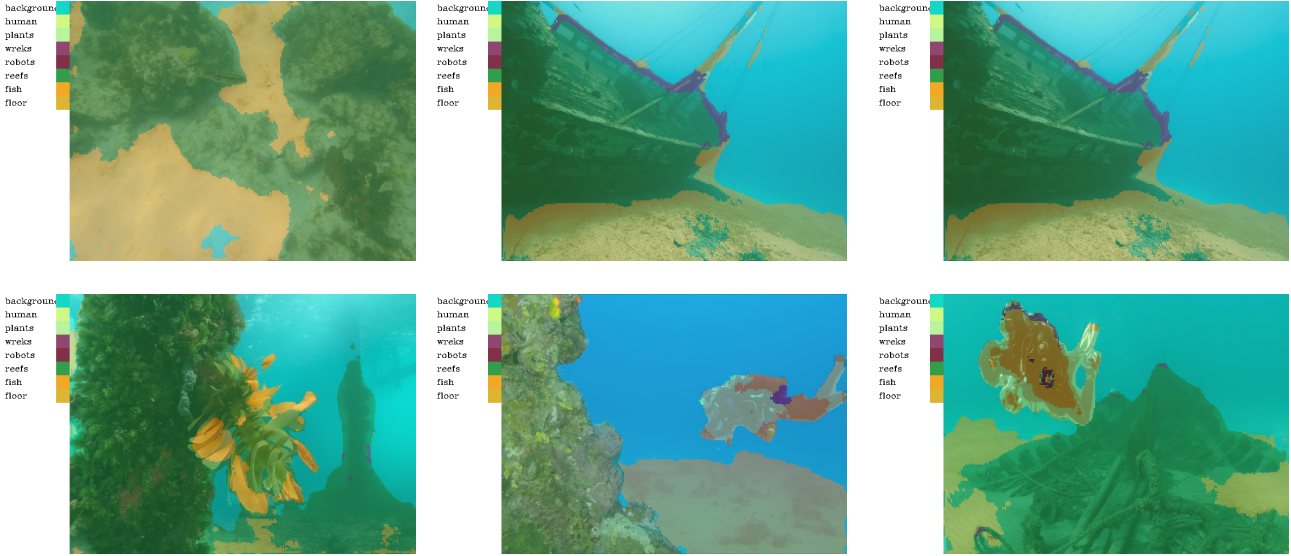

Figure 3. Qualitative results on the SUIM dataset

**VGG UNet :** A powerful semantic segmentation network, leveraging VGG's [9] deep and expressive features for robust segmentation. Offers high-quality segmentation results at the expense of increased computational complexity.

**ResNet50 UNet :** ResNet50's [3] deep residual blocks for highly accurate and detailed segmentation results. Balances between computational efficiency and superior performance, making it suitable for various segmentation tasks.

**PSP Pretrained :** In this we use a pre-trained PSPNet model on the ADE 20K dataset. Here the model is pretrained specifically on the semantic segmentation task.

## 4. Results

The results for CamVid, sitting people and SUIM datasets are in Table 1, 2 and 3 respectively. For CamVid and sitting datasets, the pretrained PSPNet yields the best mIoU scores. For the SUIM dataset, MobileNet UNet gives the best results. The qualitative visualizations of the best models are in Figure 1, 2 and 3.

## 5. Conclusion

Our paper introduces a comprehensive library for semantic segmentation for well-known models such as SegNet, FCN, UNet, and PSPNet. The library empowers researchers and practitioners in the field of computer vision with a toolset to achieve pixel-level understanding of images. We have demonstrated the efficacy and robustness of these models, underscoring their applicability in addressing diverse segmentation applications.

## 6. Acknowledgements

We would like to thank all the contributors from the open-source community. We would also like to thank ChatGPT which helped in the writing of this paper.